\documentclass[manuscript,screen, final]{acmart}
\AtBeginDocument{%
  }

\usepackage[T1]{fontenc}
\usepackage[utf8]{inputenc}
\usepackage{tcolorbox}
\tcbuselibrary{breakable}
\usepackage{placeins}      
\usepackage{tikz}

\usepackage{caption} 
\setcopyright{acmlicensed}
\copyrightyear{2025}
\acmYear{2025}
\acmDOI{XXXXXXX.XXXXXXX}
\acmConference[MM'25]{33rd ACM International Conference on Multimedia}{October 27--31,
  2025}{Dublin, Ireland}
\acmISBN{978-1-4503-XXXX-X/2018/06}

\begin{document}

\title{Analyze-Prompt-Reason: A Collaborative Agent-Based Framework for Multi-Image Vision-Language Reasoning}


\author{Angelos Vlachos}
\email{aavlachos@cslab.ece.ntua.gr}
\orcid{0009-0000-9467-6754}

\author{Giorgos Filandrianos}
\email{geofila@islab.ntua.gr}
\orcid{0000-0002-7015-7746}

\author{Maria Lymperaiou}
\email{marialymp@islab.ntua.gr}
\orcid{0000-0001-9442-4186}

\author{Nikolaos Spanos}
\email{nspanos@ails.ece.ntua.gr}
\orcid{0009-0001-2691-0956}

\author{Ilias Mitsouras}
\email{iliasmits@ails.ece.ntua.gr}
\orcid{0009-0006-3918-2616}

\author{Vasileios Karampinis}
\email{vkarampinis@ails.ece.ntua.gr}
\orcid{0009-0005-4839-6192}

\author{Athanasios Voulodimos}
\email{thanosv@mail.ntua.gr}
\orcid{0000-0002-0632-9769}

\affiliation{%
  \institution{Artificial Intelligence and Learning Systems Laboratory,
               National Technical University of Athens} 
  \city{Athens}
  \country{Greece}
}
\renewcommand{\shortauthors}{Vlachos et al.}

\begin{abstract}
We present a Collaborative Agent-Based Framework for Multi-Image Reasoning. Our approach tackles the challenge of interleaved multimodal reasoning across diverse datasets and task formats by employing a dual-agent system: a language-based \textit{PromptEngineer}, which generates context-aware, task-specific prompts, and a \textit{VisionReasoner}, a large vision-language model (LVLM) responsible for final inference. The framework is fully automated, modular, and training-free, enabling generalization across classification, question answering, and free-form generation tasks involving one or multiple input images.
We evaluate our method on 18 diverse datasets from the 2025 MIRAGE Challenge (Track A), covering a broad spectrum of visual reasoning tasks including document QA, visual comparison, dialogue-based understanding, and scene-level inference. Our results demonstrate that LVLMs can effectively reason over multiple images when guided by informative prompts. Notably, Claude 3.7 achieves near-ceiling performance on challenging tasks such as TQA (99.13\% accuracy), DocVQA (96.87\%), and MMCoQA (75.28 ROUGE-L). We also explore how design choices—such as model selection, shot count, and input length—influence the reasoning performance of different LVLMs.

\end{abstract}


\received{10 July 2025}
\received[accepted]{31 July 2025}

\maketitle

\section{Introduction}

The increasing demand for intelligent systems capable of understanding and reasoning over complex multimodal inputs has led to the emergence of benchmarks that push the limits of current vision-language models \cite{xu2024lvlm, suzuki2025resampling, bitton2023visit, chen2024we}. While reasoning with large language models (LLMs) has shown strong performance in structured textual tasks \cite{plaat2024reasoning, giadikiaroglou-etal-2024-puzzle}, extending these capabilities to complex visual reasoning, particularly in multi-image settings, remains a significant challenge \cite{kazemi2024remi, liu2024mibench}. Multi-image comprehension introduces unique demands: it requires not only visual grounding, but also the ability to integrate information across visual instances, maintain cross-modal consistency, and generate answers that align with diverse task-specific formats \cite{kazemi2024remi, zhao2024benchmarking, liu2024mibench, wahed2024prima}. 


The MM25 Grand Challenge on Multimodal Interleaved Reasoning and Generation (MIRAGE)\footnote{\url{https://mm25mirage.github.io/mirage/}} represents a significant step in this direction. Specifically, Track A focuses on Multimodal Interleaved Instruction Reasoning and aims to evaluate analytical, inferential, and comparative reasoning across a diverse set of tasks, including Multi-Image Reasoning, Document and Knowledge-Based Understanding, Interactive Multi-Modal Communication, and Multi-Image Discrimination.

In this paper, we introduce a general-purpose collaborative agent-based framework for tackling this challenge. The core idea is that each diverse task requires a well-designed prompt to properly evaluate the performance of the tested Large Vision-Language Models (LVLMs). However, when studying a new task, understanding its key aspects and difficulties is inherently challenging, which in turn hinders the generalizability of existing methods and makes the automatic adaptation and evaluation of LVLMs on new tasks practically difficult. On the other hand, using an overly simple instruction to seek an answer to a complex question requiring advanced reasoning (e.g., \textit{"What are the differences between the two birds?"}), especially when domain expertise is needed or the response format is constrained, is ineffective and obscures the actual reasoning performance of the LVLM.

Thus, inspired by techniques such as \citep{yanglarge}, we propose \textbf{Analyze-Prompt-Reason}, a dual-agent approach for multimodal reasoning. Our method consists of two main components: an LLM that assumes the role of a \textbf{PromptEngineer}, and a LVLM that acts as the \textbf{VisionReasoner}. Crucially, our system requires \textit{no} task-specific fine-tuning or human supervision. Instead, it leverages few-shot prompting and a collaborative agent strategy to autonomously construct and execute prompts tailored to the specific demands of each task.

By using this method, we seek to answer \textit{to what extent can an LVLM with task-specific prompts solve difficult multi-image tasks out of the box?} 
The key insight behind our approach lies in decoupling the problem into two synergistic steps: (1) prompt generation via an LLM that is aware of task semantics, dataset structure, and answer format, and (2) visual-textual reasoning via a general-purpose LVLM operating under the guidance of the generated prompts. Evaluated on 18 diverse datasets from the MIRAGE Track A challenge, our method demonstrates strong adaptability and performance across classification, QA, and generation settings. Despite its simplicity and generality, the framework delivers competitive results, underscoring the promise of modular, prompt-driven architectures for scalable multimodal reasoning.

\section{MIRAGE Challenge}
The MIRAGE Challenge, part of the MM25 Grand Challenge series, evaluates multimodal reasoning and generation via two tracks: (A) \textit{Multimodal Interleaved Instruction Reasoning} and (B) \textit{Multimodal Interleaved Content Generation}. This paper focuses on Track A, which tests models’ ability to follow instructions and reason over interleaved image-text sequences, integrating multiple visual contexts to produce coherent answers to open-ended or multiple-choice questions. The next section details the datasets and task categories in Track A.

\subsection{Dataset}

Track A of the MIRAGE Challenge evaluates the instruction-following and reasoning capabilities of vision-language systems across a diverse set of tasks involving image-text interleaving. These tasks are organized into four core subcategories, each focusing on a distinct aspect of multimodal reasoning. Table~\ref{tab:mirage_dataset_summary} summarizes these subcategories, providing a brief description of each and listing the associated datasets.

\begin{table}[!ht]
\centering
\small
\begin{tabular}{p{3cm} p{4cm} p{7cm}}
\toprule
\textbf{Subcategory} & \textbf{Focus} & \textbf{Datasets} \\
\midrule
\textbf{Multi-Image Reasoning} & Reasoning over multiple related images (e.g., change detection, visual entailment, fine-grained comparison) & 
Spot-the-Diff\cite{spotthediff}, CLEVR-Change\cite{clevr}, IEdit\cite{iedit}, Birds-to-Words\cite{birds}, nuScenes\cite{nuscenes}, VISION\cite{vision}, Fashion200K\cite{fashion}, MIT-States (Property/State)\cite{mit}, RecipeQA-ImageCoherence\cite{recipeqa}, NLVR2\cite{nlvr2}, VizWiz\cite{vizwiz} \\
\midrule
\textbf{Document and Knowledge-Based Understanding} & 
Understanding structured visual content and integrating external knowledge (e.g., OCR, layout, factual QA) & 
SlideVQA\cite{slivevqa}, OCR-VQA\cite{ocr}, DocVQA\cite{docvqa}, WebQA\cite{webqa}, TQA\cite{tqa}, MMQA\cite{mmqa} \\
\midrule
\textbf{Interactive Multi-Modal Communication} & Dialogue and instruction-following involving image grounding and temporal context & 
ALFRED\cite{alfred}, MMCoQA\cite{mmcqa} \\
\midrule
\textbf{Multi-Image Discrimination} & Visual similarity, identity, and differentiation across image pairs & 
Totally-Looks-Like\cite{totallylookslike}, LFW\cite{lfw} \\
\bottomrule
\end{tabular}
\caption{Overview of the four MIRAGE subcategories and their associated datasets.}
\label{tab:mirage_dataset_summary}
\end{table}

\subsection{Metrics}

Track A tasks are evaluated using two primary metrics, depending on the task format: \textbf{Accuracy} for classification-style and multiple-choice QA tasks, and \textbf{ROUGE-L} for open-ended generative tasks. More details about the metric definitions and implementation are provided in Appendix~\ref{app:metrics}. The final score is computed as the average of all metric scores across the evaluated tasks within the track. Thus, the overall score effectively captures both the discriminative and generative capabilities of multimodal systems under the instruction reasoning framework.

\section{Methodology}

\begin{figure}[ht]
    \centering
    \includegraphics[width=0.9\textwidth]{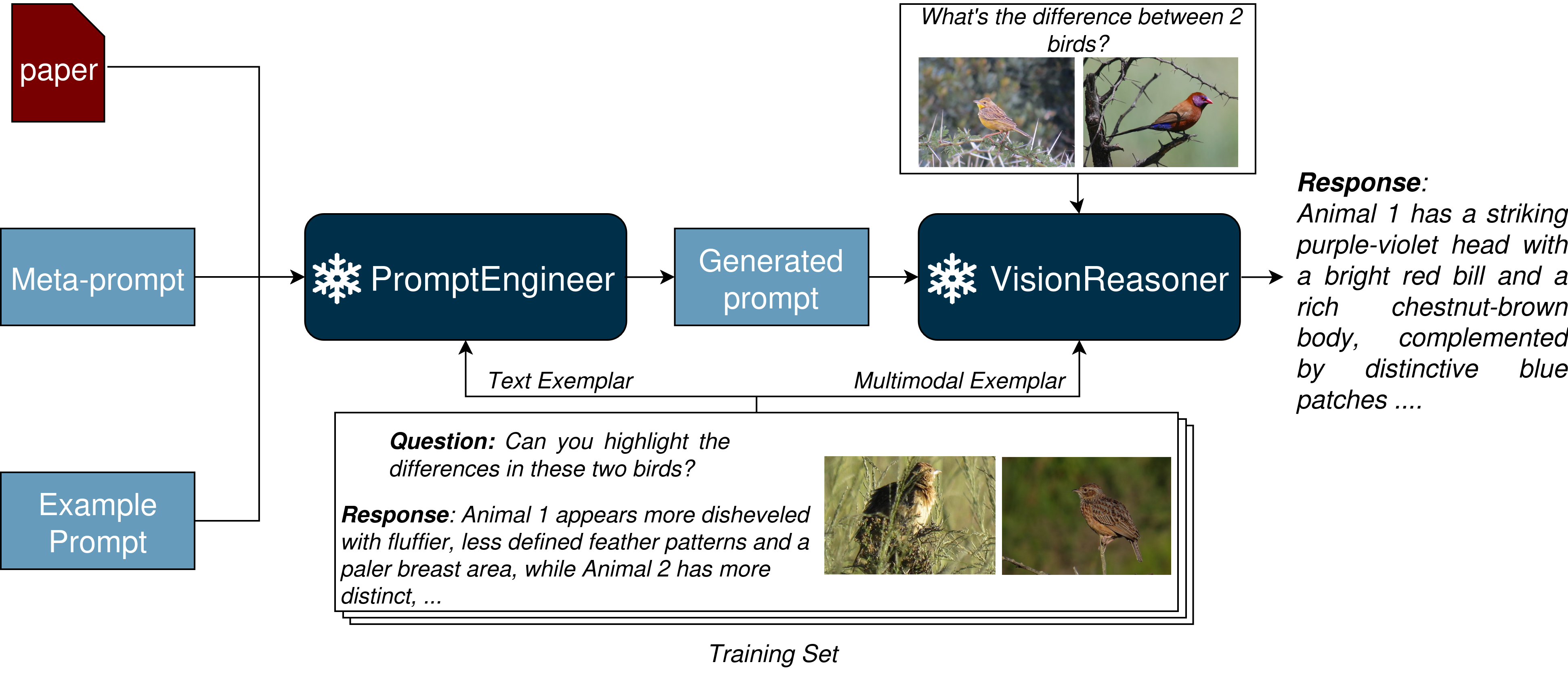} 
    \caption{Overview of the system architecture.}
    \label{fig:system-architecture}
\end{figure}

We propose an end-to-end, plug-and-play framework for Multi-Image Reasoning that operates without any task-specific training, supervision, or fine-tuning.
The method is grounded in few-shot prompting and leverages the coordinated collaboration between an LLM and an LVLM. These two models are organized into a dual-agent system, each with a distinct role: one responsible for prompt generation, and the other for executing visual reasoning. An overview of the method is shown in Figure \ref{fig:system-architecture}.

\subsection{Overview}

The \textbf{Analyze-Prompt-Reason} architecture consists of two key components:

\begin{itemize}
\item \textbf{PromptEngineer}: an LLM responsible for task \textbf{analysis} and the generation of informative \textbf{prompts} to guide the LVLM.
\item \textbf{VisionReasoner}: an LVLM that performs multi-image \textbf{reason} to produce the final output.
\end{itemize}

Each component is designed to operate independently yet cooperatively, enabling generalization across diverse tasks and datasets. 

\subsection{PromptEngineer: Prompt Generation Agent}

The \textbf{PromptEngineer} is the first agent in our framework and plays a critical meta-cognitive role. It is guided by a high-level instruction, referred to as the \textit{meta-prompt}, which defines its objective: to synthesize a coherent and informative prompt tailored to the capabilities of the VisionReasoner. This meta-prompt includes the following elements:

\begin{enumerate}
\item A detailed task definition describing the role of the PromptEngineer.
\item The research paper or accompanying document that provides context and background on the dataset.
\item The type of task that the VisionReasoner is expected to solve (e.g., classification, multiple-choice, open-ended generation) along with a representative example question from the dataset that the VisionReasoner must answer.
\item An example of a manually created prompt for a similar task which acts as the prompt prototype.
\item A few-shot set of desired output examples (text-only) from the training set, illustrating the expected structure and level of detail in the answers.
\end{enumerate}

Upon receiving this information, the PromptEngineer analyzes the provided material to extract relevant domain knowledge, understand the dataset’s construction and reasoning requirements, and internalize the target output format. It then generates a detailed prompt aimed at instructing the VisionReasoner in a way that preserves both the semantic fidelity of the task and the expected answer structure. This stage is crucial, as it effectively translates abstract task requirements into actionable input for the LVLM.

\subsection{VisionReasoner: Reasoning Agent}

The second agent, \textbf{VisionReasoner}, is a vision-language model tasked with performing the actual reasoning and generating the final answer. It receives three inputs:

\begin{enumerate}
\item The prompt generated by the PromptEngineer, which includes all necessary task instructions.
\item A few-shot set of paired image-inputs and textual answers from the dataset, allowing the model to infer the task format and content distribution.
\item A new input instance for which the model is required to generate the final answer.
\end{enumerate}

Given these inputs, the VisionReasoner performs integrated reasoning over visual and textual modalities, relying solely on the guidance provided by the prompt and the multimodal few-shot examples drawn from the training dataset. These few-shot examples help the model empirically understand the task format, the required depth of reasoning, and the expected output structure. By observing input-output patterns from a small set of representative instances, the model can better generalize to unseen questions and align its responses with the intended semantics of the task.


The goal of our experiments is not to introduce a novel approach for multi-image reasoning, but rather to investigate the extent to which a simple multi-agent framework can, on its own, address such heterogeneous tasks without any human supervision.

\section{Experiments}


\subsection{Experimental Setup}

\textit{\textbf{Validation Split.}} To monitor performance during prompt engineering and ablation studies, we carve out a dedicated \emph{validation} subset from each dataset’s original training partition whose cardinality is exactly three times that of the official test split (1500 samples). The resulting pool provides a stable basis for hyperparameter exploration and prompt selection. We note, however, that for the \textsc{MMQA} and \textsc{WebQA} datasets, the training sets do not include the full set of multiple-choice options per question; as such, these datasets are excluded from our validation procedure.

\textit{\textbf{Models.}} All prompts are auto-generated with \texttt{GPT-4o}\footnote{gpt-4o-2024-11-20} to guarantee consistent, high-quality query formulations across datasets.
For answer generation we evaluate two state-of-the-art LVLMs: \emph{Claude Sonnet 3.5}\footnote{anthropic.claude-3-7-sonnet-20250219-v1:0} and \emph{Claude Sonnet 3.7}\footnote{anthropic.claude-3-5-sonnet-20241022-v2:0}.
These models are widely adopted in industry and academia, making them ideal baselines for the MIRAGE Challenge.

\textit{\textbf{Prompting Strategy.}} We adopt few-shot, \textit{in-context} learning with mixed textual–visual exemplars. For datasets with consistently two images per instance (e.\,g., \texttt{Spot-the-Diff}, \texttt{NLVR2}, \texttt{VizWiz}), we include up to three exemplars (3-shot). For datasets with higher or more variable per-instance image counts, we reduce the number of shots accordingly. More details about the prompts used can be found in Appendix~\ref{app:prompts}. 

\begin{table}[!ht]
\centering
\small
\begin{tabular}{l|cccc|cccc}
\toprule
\textbf{Dataset} & \multicolumn{4}{c|}{\textbf{Claude 3.5}} & \multicolumn{4}{c}{\textbf{Claude 3.7}} \\
 & \textbf{0-shot} & \textbf{1-shot} & \textbf{2-shot} & \textbf{3-shot} & \textbf{0-shot} & \textbf{1-shot} & \textbf{2-shot} & \textbf{3-shot} \\
\midrule
\multicolumn{9}{c}{\textbf{ROUGE-L}} \\
\midrule
Spot-the-Diff & 12.41 & 13.05 & 14.66 & 13.58 & 16.85 & 16.85 & 18.69 & 17.01 \\
CLEVR-Change & 12.87 & 17.63 & 12.23 & 19.82 & 17.47 & 25.32 & 26.71 & 28.46 \\
IEdit & 16.02 & 14.21 & 15.34 & 14.80 & 20.26 & 17.76 & 18.27 & 17.34 \\
Birds-to-Words & 12.05 & 11.71 & 13.07 & 13.56 & 12.50 & 12.95 & 12.92 & 13.47 \\
ALFRED & 36.95 & 38.46 & 39.60 & 41.00 & 37.27 & 36.94 & 38.64 & 38.52 \\
MMCoQA & 64.30 & 69.58 & 71.55 & 71.29 & 70.64 & 73.55 & 75.16 & 75.28 \\
\midrule
Average - ROUGE-L & 25.77 & 27.44 & 27.74 & 29.01 & 29.17 & 30.56 & \textbf{31.73} & 31.68 \\
\midrule
\multicolumn{9}{c}{\textbf{Accuracy}} \\
\midrule
nuScenes & 50.33 & 53.27 & 51.4 & - & 56.00 & 60.00 & 59.93 & - \\
VISION & 91.47 & 92.13 & 91.93 & 90.4 & 84.67 & 84.80 & 88.67 & 88.4 \\
Fashion200K & 15.27 & 22.27 & 26.73 & - & 9.80 & 16.33 & 15.20 & - \\
MIT-States\_PropertyCoherence & 70.33 & 74.93 & 74.67 & 75.93 & 70.40 & 74.27 & 74.20 & 75.73 \\
MIT-States\_StateCoherence & 53.67 & 56.07 & 54.73 & 56.07 & 52.53 & 53.47 & 54.87 & 56.53 \\
RecipeQA\_ImageCoherence & 76.80 & 85.13 & - & - & 79.20 & 91.87 & - & - \\
NLVR2 & 99.80 & 99.93 & 99.93 & 99.93 & 99.93 & 99.93 & 99.93 & 99.93 \\
VizWiz & 31.00 & 41.0 & 39.53 & 38.53 & 30.27 & 46.47 & 40.73 & 53.93 \\
SlideVQA & 88.87 & 89.73 & 89.87 & 89.87 & 85.53 & 87.60 & 89.67 & 90.07 \\
OCR-VQA & 49.87 & 53.6 & 65.47 & 61.33 & 51.27 & 62.27 & 64.93 & 67.20 \\
DocVQA & 92.27 & 93.27 & 93.07 & 94.07 & 84.47 & 87.07 & 87.80 & 96.87 \\
TQA & 70.60 & 70.8 & 71.4 & 98.80 & 68.53 & 70.53 & 72.80 & 99.13 \\
\midrule
Average - Accuracy & 65.86 & 69.34 & 68.98 & 78.33 & 64.38 & 69.55 & 68.07 & \textbf{80.87} \\
\midrule
Overall Average & 45.82 & 48.39 & 48.36 & 53.67 & 46.78 & 50.06 & 49.9 & \textbf{56.28} \\
\bottomrule
\end{tabular}
\caption{Performance of Claude 3.5 and 3.7 on MIRAGE Challenge validation sets with varying shot counts in VisionReasoner. Entries marked with `--' were skipped due to the increased number of images per example, exceeding the LVLM’s context length.}

\label{tab:mirage_claude_unified}
\end{table}
\subsection{Results}

Table~\ref{tab:mirage_claude_unified} presents the performance of Claude Sonnet 3.5 and 3.7 on the MIRAGE Challenge validation split across a diverse set of vision-language tasks. The table is structured by evaluation metric—ROUGE-L for captioning and generative subtasks, and Accuracy for classification and QA subtasks. For each model, results are reported under 1-shot, 2-shot, and 3-shot settings using the VisionReasoner prompting framework.

\textit{\textbf{What is the impact of model selection?}} Choosing between Claude 3.5 and 3.7 yields only modest gains overall—Claude 3.7 edges out 3.5 on average ROUGE‑L (29.17→31.73 at 2‑shot) and Accuracy (64.38→80.87 at 3‑shot), yet the gap varies by task and shot count. In zero‑shot settings 3.7 outperforms 3.5 (46.78 vs. 45.82 overall), but on some datasets (e.g.\ ALFRED, Birds‑to‑Words, Fashion200K) 3.5 still leads. Thus, while 3.7 is generally preferable, task‑specific characteristics can make 3.5 the better choice.

\textit{\textbf{How does the number of shots affect performance?}} To evaluate how the number of few-shot examples influences performance, we compare 0-shot, 1-shot, 2-shot, and 3-shot variants across all tasks. In several cases, such as TQA and DocVQA, increased shot counts lead to incremental performance gains, suggesting that the additional examples help the model better capture procedural or document-level reasoning patterns. However, in other tasks like Fashion200K or CLEVR-Change, performance does not monotonically improve with more shots. This suggests that prompt quality and representativeness matter more than sheer quantity, and that non-monotonic trends may result from noise introduced by suboptimal examples or context window pressure. There are also datasets like TQA where the number of shots dramatically boosts accuracy, for example, from 68.53\% in the 0-shot setting to 99.13\% in the 3-shot case.


\textit{\textbf{Can LVLMs solve multimodal interleaved instruction reasoning tasks?}} Perhaps the most striking result is the strong performance of VisionReasoner in the complete absence of human supervision. Tasks such as DocVQA (96.87\%), TQA(99.13\%) and MMCoQA (75.28 ROUGE-L) are solved at near-ceiling levels using only automatically selected few-shot examples and simple task prompts. This confirms that general-purpose, fully automatic prompting pipelines can match or exceed the performance of hand-tuned approaches in well-defined tasks. These findings suggest that, for many applications, labor-intensive dataset-specific pipelines can be replaced by unified prompting strategies that scale more naturally across domains.

\section{Conclusion}
We present \textbf{Analyze-Prompt-Reason}, an automated, agent-based framework to assess whether LVLMs can solve complex multi-image reasoning tasks without supervision. Our dual-agent setup—combining task-aware \textbf{Prompt} generation with visual \textbf{Reason}ing—achieves strong performance across diverse MIRAGE Track A tasks. Notably, models like Claude 3.7 reach near-perfect accuracy on benchmarks such as TQA (99.13\%) and DocVQA (96.87\%), using only few-shot, auto-generated prompts. These results highlight the potential of unified prompting pipelines to replace hand-crafted, task-specific solutions. Future work will focus on optimizing shot selection and extending to broader model families.


\begin{acks}
We acknowledge the use of Amazon Web Services (AWS), for providing the cloud computing infrastructure that enabled the deployment and use of the large language models (LLMs) utilized in this study.
\end{acks}

\bibliographystyle{ACM-Reference-Format}
\bibliography{sample-manuscript}

\appendix

\section{Metrics}
\label{app:metrics}

The primary evaluation metrics adopted in Track A of the MIRAGE Challenge are presented below.

\paragraph{\textbf{Accuracy}}

Accuracy is used to measure the correctness of predicted answers in tasks that involve discrete choice options. It is computed as:

\[
\text{Accuracy} = \frac{1}{N} \sum_{i=1}^{N} \mathbb{1}\left[\hat{y}_i = y_i\right]
\]

where $\hat{y}_i$ is the predicted answer for instance $i$, $y_i$ is the ground-truth answer, and $\mathbb{1}[\cdot]$ is the indicator function.

\paragraph{\textbf{ROUGE-L}}
This metric is used for evaluating the quality of generated textual responses by comparing the longest common subsequence (LCS) between the model output and a reference text. The F1 variant balances precision and recall:

\[
\text{ROUGE-L}_{F1} = \frac{(1 + \beta^2) \cdot P \cdot R}{P + \beta^2 \cdot R}
\]

where $P$ and $R$ are the LCS-based precision and recall, respectively, and $\beta$ is typically set to 1.

\section{Prompts}
\label{app:prompts}
In our methodology, we used two main prompts—one for each component, namely PromptEngineer and VisionReasoner. The prompts for these components are as follows.

\FloatBarrier            

\begin{tcolorbox}[colback=gray!5!white,
                  colframe=black!75!black,
                  title=PromptEngineer's Prompt (Meta-prompt),
                  fonttitle=\bfseries,
                  sharp corners=south,
                  breakable]    
\footnotesize
\begin{verbatim}
You are the PromptEngineer agent. Your single responsibility is to craft a high-quality few-shot prompt that prepares the
VisionReasoner model to solve tasks drawn from the target dataset—without any extra commentary.

---
INPUT PACKAGE
---
1. <DATASET_PAPER>       – full paper describing the dataset’s goals, collection protocol, and annotation scheme
2. <TASK_TYPE>           – e.g. classification, multiple-choice, open generation
3. <REPRESENTATIVE_Q>    – one typical question from the dataset
4. <EXAMPLE_PROMPT>      – a hand-written prompt for a *different* dataset (use this only as a stylistic reference)
5. <FEW_SHOT_EXAMPLES>   – text-only QA-pairs showing the exact answer structure the VisionReasoner must reproduce

---
MANDATED WORKFLOW
---
A. Study <DATASET_PAPER>
   • Extract essential domain knowledge, key entities, and reasoning patterns
   • Note any dataset-specific instructions, constraints, or evaluation metrics

B. Analyze <EXAMPLE_PROMPT> and <FEW_SHOT_EXAMPLES>
   • Internalize tone, concision, and structural conventions
   • Identify required answer fields, ordering, and formatting cues

C. Draft the VisionReasoner Prompt
   Your prompt MUST:
   • Start with a concise task definition
   • Summarise critical background gleaned from the paper (max. 3 sentences)
   • Provide clear, numbered instructions for the model
   • Include placeholders (e.g. {question}, {choices}) for dynamic content
   • Supply the <FEW_SHOT_EXAMPLES> verbatim in a "### Examples" block
   • End with "### Now answer:" to signal the model to respond

D. Validate
   • Ensure the draft is self-contained—VisionReasoner should not need any outside context beyond what you embed
   • Match the exact answer formatting visible in <FEW_SHOT_EXAMPLES>
   • Remove all explanatory comments or meta-notes

---
Reference prompt and few shot examples
---
### Reference Prompt 
<EXAMPLE_PROMPT>

### Examples
<FEW_SHOT_EXAMPLES>

---
OUTPUT
---
Return *only* the final prompt text, ready for direct use. Do NOT prepend or append explanations, markdown fences, or LaTeX
commands.
\end{verbatim}

\end{tcolorbox}
\FloatBarrier

\FloatBarrier  

\begin{tcolorbox}[colback=gray!5!white,
                  colframe=black!75!black,
                  title=VisionReasoner's Prompt,
                  fonttitle=\bfseries,
                  sharp corners=south,
                  breakable]

\footnotesize
\begin{verbatim}
<PROMPT GENERATED FROM PROMPT ENGINENER>

---

To help you understand the format and the task, I will provide you with {num_examples_text}. Do not include any additional 
information, context, or explanations—only the differences, strictly following the format.

<FEW_SHOT_EXAMPLES>
---
Test instance

<test instance>
\end{verbatim}

\end{tcolorbox}

The <test instance> within the VisionReasoner’s prompt represents the input for which a response is to be generated.

All exemplars are randomly drawn from the training set and kept \emph{fixed} across runs to eliminate variance from exemplar selection. For the PromptEngineer, we also include an example prompt, a manually crafted instance from the DocVQA dataset, used solely to guide prompt generation. VisionReasoner, however, relies on the generated version. The manually created example prompt is presented below.

\FloatBarrier  

\FloatBarrier  

\begin{tcolorbox}[colback=gray!5!white,
                  colframe=black!75!black,
                  title=Example Prompt (DocVQA),
                  fonttitle=\bfseries,
                  sharp corners=south,
                  breakable]

\footnotesize
\begin{verbatim}
You are a document understanding assistant. Your task is to read one or more document images and answer a question based on
the information presented in them.
 
Each task includes a small set of images (usually 1 to 6 scanned documents), a natural-language question, and a list of
possible answer choices. Your job is to find the correct answer using only the information found in the images.
 
Here’s how you should approach the task:
     1.  Read all the provided images carefully. These documents may include forms, tables, invoices, letters, memos, 
     balance sheets, or other structured or unstructured formats.
     2.  Look for information in both printed and handwritten text. Important content might appear in tables, titles, 
     headers, footnotes, or embedded within the layout.
     3.  Once you’ve found the relevant information, choose the correct answer from the list of choices.
     4.  Your answer should be copied exactly from the choice list — spelling, punctuation, and formatting must match 
     perfectly.
     5.  Do not include any extra words, explanations, or punctuation. Just return the selected answer.
 
This task is evaluated using strict accuracy metrics, so even small differences in formatting (like an extra space or 
missing punctuation) can cause your answer to be marked wrong.
 
 
Remember: look carefully at the images, match your answer exactly to one of the given options, and do not include anything 
else in your output.
To help you understand the format and the task, I will provide you with {num_examples_text}. Do not include any additional 
information, context, or explanations—only the differences, strictly following the format.

<FEW_SHOT_EXAMPLES>

\end{verbatim}

\end{tcolorbox}

\FloatBarrier  

\end{document}